\documentclass[letterpaper]{article} 
\usepackage{aaai24}  
\usepackage{times}  
\usepackage{helvet}  
\usepackage{courier}  
\usepackage[hyphens]{url}  
\usepackage{graphicx} 
\usepackage{natbib}  
\usepackage{caption} 
\usepackage{algorithm}
\usepackage{algorithmic}

\usepackage{newfloat}
\usepackage{listings}
\DeclareCaptionStyle{ruled}{labelfont=normalfont,labelsep=colon,strut=off} 
\floatstyle{ruled}
\newfloat{listing}{tb}{lst}{}
\floatname{listing}{Listing}
\usepackage{booktabs}
\usepackage{subcaption}

\title{Minding Motivation: The Effect of Intrinsic Motivation on Agent Behaviors}
\author{
    Leonardo Villalobos-Arias\thanks{Thanks to the University of Costa Rica for their support.},
    Grant Forbes, Jianxun Wang, David L Roberts, Arnav Jhala
}
\affiliations{
    North Carolina State University\\

    Department of Computer Science\\
    Raleigh, North Carolina, USA\\
    \{lvillal, gforbes, jwang75, dlrober4, ahjhala\}@ncsu.edu
}

\begin{document}

\maketitle

\begin{abstract}
Games are challenging for Reinforcement Learning~(RL) agents due to their reward-sparsity, as rewards are only obtainable after long sequences of deliberate actions. Intrinsic Motivation~(IM) methods---which introduce exploration rewards---are an effective solution to reward-sparsity. However, IM also causes an issue known as `reward hacking' where the agent optimizes for the new reward at the expense of properly playing the game. The larger problem is that reward hacking itself is largely unknown; there is no answer to whether, and to what extent, IM rewards change the behavior of RL agents. This study takes a first step by empirically evaluating the impact on behavior of three IM techniques on the MiniGrid game-like environment. We compare these IM models with Generalized Reward Matching~(GRM), a method that can be used with any intrinsic reward function to guarantee optimality. Our results suggest that IM causes noticeable change by increasing the initial rewards, but also altering the way the agent plays; and that GRM mitigated reward hacking in some scenarios.
\end{abstract}

\section{Introduction}

The phrase ``limitation breeds creativity'' has never been more applicable to artificial intelligence in video games than now. Some of the most important breakthroughs in Reinforcement Learning~(RL) research have been reached through game-playing agents~\cite{mnih2013playing,mnih2015human,schulman_proximal_2017,vinyals2019grandmaster}. Reward-sparse games are characteristically difficult for RL to learn due to the long sequence of actions required to both discover and then properly attribute sparse rewards~\cite{pathak_curiosity-driven_2017,huang_action_2020}\footnote{The advances in reinforcement learning methods and environments such as MiniGrid and MicroRTS provided a strong foundation for this work to be possible.}. Research on Intrinsic Motivation~(IM)---a method that enhances the environment with exploration rewards---has led to RL agents that can perform well on these hard games~\cite{burda_large-scale_2018}.

Intrinsic Motivation, while useful and even necessary, has its own set of drawbacks. IM is prone to `reward hacking', a phenomena where the agent optimizes for the shaped reward at the expense of the actual reward~\cite{forbes_potential-based_2024}. As~\citet{burda_exploration_2018} describe it, their IM agents position themselves next to hazards---a behavior aptly named ``dancing with skulls''---for the intrinsic reward associated with their inherent rarity at the expense of playing the game. A related issue is the `noisy-TV' problem~\cite{burda_large-scale_2018}, where the agent, distracted by intrinsic rewards, disregards the search of the real reward. Ongoing research on policy-invariant methods harnesses the benefits of IM while reducing its negative effects~\cite{raileanu2020ride,behboudian_policy_2022,forbes_potential-based_2024}.

A larger problem is that the effects of Intrinsic Motivation on the way RL agents behave are largely unknown. The state-of-the-art experiment designs~\cite{taiga2020bonus,andres2022evaluation,forbes_potential-based_2024} focus only on the maximization of rewards over time. \citet{burda_exploration_2018} discovered an instance of the reward hacking bug by visualizing its policy, yet we are only aware of a handful of papers that actively check the effect of IM using such methods~\cite{huang_action_2020,le_beyond_2024,raileanu2020ride,kayal_impact_2025}. Moreover, experiments are often executed on environments where RL can not properly learn without IM. With no baseline behavior, evaluating the policy-invariance of IM becomes impossible. For these reasons, there is no answer to what extent, and under which conditions, intrinsic rewards affect the final behavior of game-playing agents.

To address this gap in literature, we propose an empirical evaluation on the effect of intrinsic motivation techniques on the behavior of reinforcement learning agents. We measure the impact of these methods in terms of both reward performance and exhibited behavior. We base the protocol of this study on \cite{kayal_impact_2025}, and select three traditional IM methods to evaluate: State Counting, Max Entropy, and Intrinsic Curiosity Model~(ICM). As a representative of policy-invariant methods, we selected Generalized Reward Matching~(GRM, specifically D-GRM)~\cite{forbes_potential-based_2024} and combine it with these three IM sources. We train the agents on Minigrid~\cite{chevalier-boisvert_minigrid_2023}, a simplification of game-like environments, since it allows behavioral analysis in the form of policy visualization. An important distinction with previous IM evaluations is our conscious selection of environments can be learned with no-IM motivation. To talk about behavior in RL,and how different IM methods result on different behaviors, it is a necessity to have a baseline policy. Our main contribution is an empirical analysis on the policy variance of various IM techniques. The results of this experiment also double as a benchmark of GRM, which previously only was tested on Montezuma's Revenge~\cite{forbes_potential-based_2024}.

\section{Related Work}

Intrinsic Motivation is a subfield of reward shaping that augments a sparse-reward environment with an additional reward function based on ``intrinsic'' goals, which are often complex, non-Markovian shaping rewards meant to generalize well across environments, rather than being tailored to a particular environment. These rewards are often based on psychological concepts~\cite{oudeyer2007intrinsic} such as curiosity, empowerment, novelty, or skill-learning~\cite{burda_large-scale_2018, mohamed2015variational, colas2022autotelic}. Examples of intrinsic reward functions include Random Network Distillation~(RND)~\cite{burda_exploration_2018}, Intrinsic Curiosity Module~(ICM)~\cite{pathak_curiosity-driven_2017}, Discriminative-model-based Episodic Intrinsic Reward (DEIR)~\cite{wan2023deir}, Never Give Up~\cite{badia2020never}, and NovelD~\cite{zhang2021noveld}.

There is ongoing research to mitigate the side effects of intrinsic motivation. \citet{huang_action_2020} propose an algorithm called Action Guidance, which consists of learning separate policies for the real and shaped~(intrinsic) rewards, tested on the MicroRTS environment~\cite{ontanon2018first}, though it requires use of off-policy gradient methods. The EIPO algorithm proposed in \cite{chen2022redeeming} automatically scales the intrinsic reward coefficient by augmenting it when exploration is necessary, and vice-versa. \citet{le_beyond_2024} introduce the concept of ``surprise novelty'' to mitigate the noisy TV problem. \citet{raileanu2020ride} introduce Rewarding Impact-Driven Exploration~(RIDE) as a type of IM method designed for one-shot~(a state is seldom visited twice) observation problems, such as procedurally generated environments. \citet{behboudian_policy_2022} create, based on the concept of potential-based rewards, the Policy-Invariant Explicit Shaping~(PIES) algorithm that similarly diminishes the intrinsic reward to guarantee policy invariance by the end of training. Following a similar line of research, \citet{forbes2024potential, forbes_potential-based_2024,forbes2025action} extend the body of research of intrinsic motivation with three major algorithms: Potential-Based Intrinsic Motivation~(PBIM), Generalized Reward Matching~(GRM), and Action-Dependent Optimality-Preserving Shaping~(ADOPS).

Aside from the empirical analysis that often accompanies the proposal of a novel method, there is a limited number~\cite{jordan2024position} of studies within RL that benchmark existing IM methods, and particularly existing optimality-preserving IM methods. \citet{taiga2020bonus} perform a study on the Atari Learning Environment, where they rank state count, ICM, RND, and NoisyNets. \citet{laskin2021urlb} create a benchmark suite for unsupervised RL, including IM methods. \citet{andres2022evaluation} perform an empirical evaluation of intrinsic motivation hyperparameters in the MiniGrid environment. 
Lastly, \citet{kayal_impact_2025} evaluates four intrinsic motivation algorithms on the MiniGrid environment by assessing how they impact the behavior of the RL agent. 
None of these prior empirical evaluations, however, have focused on empirically benchmarking and comparing prior optimality-preserving optimality methods. In this paper, we present such an evaluation.

\section{Experimental Design}

The main objective of this evaluation is to empirically evaluating the behavior of reinforcement agents when trained using different intrinsic motivation techniques. We measure this change of behavior on two dimensions: 1) return (real/extrinsic reward) performance, and 2) exhibited policy behavior.

We base our experimental design on a recent study~\cite{kayal_impact_2025}, where the authors analyze how IM impacts exploration of RL agents. We shift focus away from emphasis on exploration and on the different levels of diversity of IM, and instead emphasize regular (policy altering) and policy-invariant IM and their impact on behavior. Our implementation is a modified version of the publicly-available DEIR method~\cite{wan2023deir}, and is publicly available on: https://github.com/lyonva/bad-apple/releases/tag/mindingmotivation. Hereon, we refer to \cite{kayal_impact_2025} as the protocol study.

\subsection{Environment}
\begin{figure}[ht]
    \centering
    \begin{subfigure}{0.15\textwidth}
        \includegraphics[width=\textwidth]{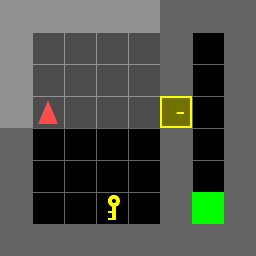}
        \caption{DoorKey-8x8}
        \label{fig:map-dk8}
    \end{subfigure}
    \hfill
    \begin{subfigure}{0.15\textwidth}
        \includegraphics[width=\textwidth]{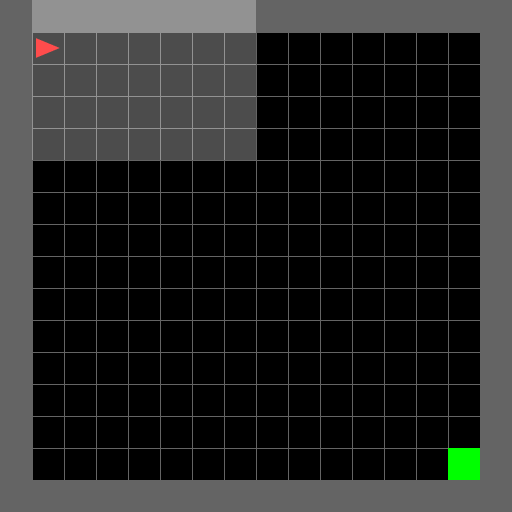}
        \caption{Empty-16x16}
        \label{fig:map-empty}
    \end{subfigure}
    \hfill
    \begin{subfigure}{0.15\textwidth}
        \includegraphics[width=\textwidth]{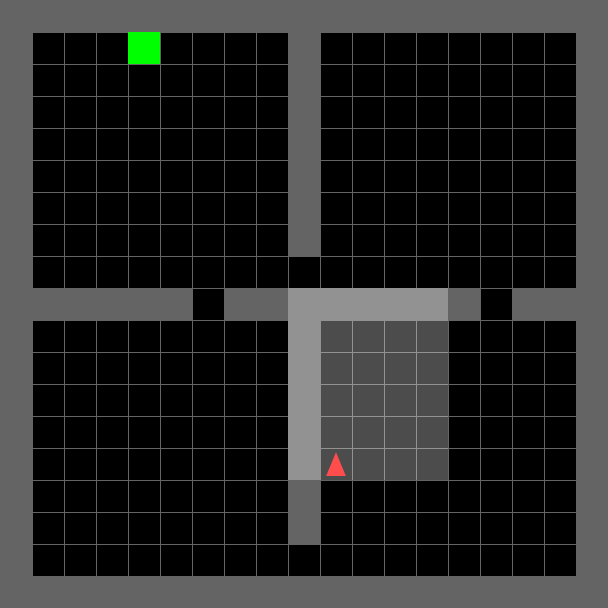}
        \caption{FourRooms}
        \label{fig:map-fr}
    \end{subfigure}

    \begin{subfigure}{0.20\textwidth}
        \includegraphics[width=\textwidth]{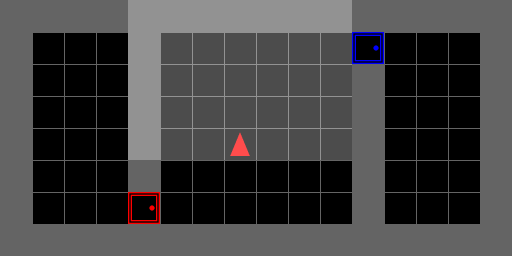}
        \caption{RedBlueDoors-8x8}
        \label{fig:map-rbd}
    \end{subfigure}
    \hfill
    \begin{subfigure}{0.15\textwidth}
        \includegraphics[width=\textwidth]{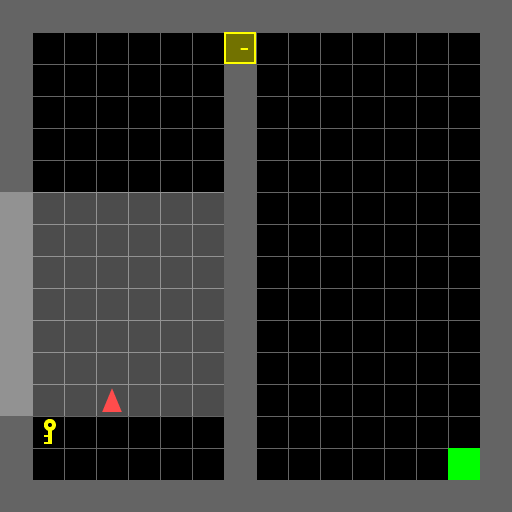}
        \caption{DoorKey-16x16}
        \label{fig:map-dk16}
    \end{subfigure}
    \hfill
    
    \caption{MiniGrid maps used in this experiment.}
    \label{fig:map}
\end{figure}

We use the same four MiniGrid~\cite{chevalier-boisvert_minigrid_2023} environments as the protocol study, with the addition of DoorKey-8x8. All of these environments are reward-sparse, as the agent only gets a non-zero reward by the end of a successful episode. The environment (extrinsic) reward is inversely proportional to the number of time steps that the agent required to get to the goal to incentivize efficiency. We train agents in the following environments, which are shown on figure~\ref{fig:map}:

\begin{itemize}
    \item \textbf{DoorKey-8x8} requires getting to the goal tile, which is in a separate room behind a locked door. The agent thus must learn to pick up a key to open the door to proceed to the goal.
    \item \textbf{Empty-16x16} has the objective of getting to the goal tile, which is always on the bottom right part of the map. The entirety of the map is empty, except for the outmost tiles which are walls.
    \item \textbf{FourRooms} features four connected rooms, where both the goal tile and the agent's initial position are randomized. Reaching the goal often requires the agent to visit and explore two or more rooms.
    \item \textbf{RedBlueDoors-8x8} features a central room with a red door on the left side and a blue door on the right side. The agent must open the doors in this order to succeed.
    \item \textbf{DoorKey-16x16} follows the same rules as DoorKey-8x8, but with a much larger space. The agent thus has a larger search space to comb through and thus rewards become much more sparser.
\end{itemize}

In contrast with the protocol study, we only use the standard grid partial observation space ($7 \times 7 \times 3$ tensor). The actions available to an agent in MiniGrid are: turn left, turn right, move forward, pickup, drop, and toggle.

\subsection{Model Architecture}
Following the protocol study, we use Proximal Policy Optimization~\cite{schulman_proximal_2017} as the base learning algorithm. In contrast to the protocol study, we have separate critic networks to calculate the extrinsic and intrinsic values. The model has a shared CNN comprised of three layers: 16 filters, 32 filters, and 64 filters; all which are sized $2 \times 2$ and use ReLU activation. This CNN outputs to the actor network and the two critics, all which have the same architecture of 64 hidden units with ReLU activation.

We train seven variants of this model: no intrinsic motivation~(baseline), 3 IM methods, and 3 IM with GRM. We exclude DIAYN due to poor performance in the protocol study:
\begin{itemize}
    \item \textbf{State Count} is a count-based method that grants a reward that is inversely proportional to the number of times a state has been observed. This intrinsic reward follows the form $1/\sqrt(N_s)$, where $N_s$ is the number of times state $s$ has been visited so far~\cite{strehl2008analysis}.
    \item \textbf{Max Entropy} awards the agent with an intrinsic reward equal to the policy network's entropy in order to incentivize stochasticity~\cite{liu2019policy}.
    \item \textbf{ICM} uses the concept of curiosity to generate intrinsic rewards~\cite{pathak_curiosity-driven_2017}. ICM relies on two auxiliary networks, for which we use a hidden layer of 256 units with ReLU activation, and a learning rate of $3e^{-4}$.
\end{itemize}

We train each model on each environment for a total of $20.48$ million frames (1,000 rollouts), and repeat this process for a total of ten runs per combination of map and IM. Table~\ref{tab:hyper} shows the selected values for the main PPO hyperparameters. In addition, we manually tune, per map and method, the hyperparameter value of the intrinsic reward coefficient $\beta$ in the range $[1, 0.5, 0.1, 0.05, 0.01, 0.005, 0.001, ..., 0.000005, 0.000001]$. Table~\ref{tab:beta} shows the final values for $\beta$.

\begin{table}[ht]
\centering
\caption{PPO hyperparameters.}
\label{tab:hyper}
\begin{tabular}{lr}
\toprule
\# parallel environments     & 16       \\
\# frames per rollout        & 128      \\
\# epochs                    & 4        \\
Batch size                   & 256      \\
Discount $\gamma$            & 0.99     \\
Learning rate                & 0.0001   \\
Entropy regularization       & 0.0005   \\
Value loss coefficient       & 0.5      \\
PPO clipping factor          & 0.2      \\
Gradient clipping            & 0.5      \\
\bottomrule
\end{tabular}
\end{table}

\begin{table*}[ht]
\centering
\caption{Chosen values for intrinsic reward coefficient $\beta$. The value is set to 0 for the no-IM model.}
\label{tab:beta}
\begin{tabular}{@{}lrrrrr@{}}
\toprule
            & DoorKey-8x8 & Empty-16x16 & FourRooms & RedBlueDoors-8x8 & DoorKey-16x16 \\
\midrule
State Count & 1           & 0.01        & 1         & 1                & 0.01          \\
Max Entropy & 0.001       & 0.00005     & 0.00001   & 0.000001         & 0.001         \\
ICM         & 0.1         & 0.01        & 0.05      & 0.000001         & 0.05          \\
GRM+SC      & 1           & 0.01        & 0.05      & 0.1              & 0.01          \\
GRM+ME      & 0.001       & 0.00005     & 0.00001   & 0.000001         & 0.001         \\
GRM+ICM     & 0.1         & 0.01        & 0.05      & 0.000001         & 0.05          \\
\bottomrule
\end{tabular}
\end{table*}

\subsection{Evaluation}
We evaluate with performance metrics and behavioral analysis with visualizations of in-environment behavior. 

We capture---per training rollout, and aggregated over all parallel actors---two evaluation metrics during training:
\begin{itemize}
    \item \textbf{Episodic return:} The average reward obtained per elapsed episode averaged across all actors. Higher is better.
    \item \textbf{Position coverage:} The percentage of unique grid positions---$(x,y)$ coordinates---visited across all actors. The total count of visited positions is divided across the number tiles that are possible to visit in each map. Higher is better.
\end{itemize}


Agents have access to their observations~($7\times7\times3$ image), but do not know their position on the grid. 

For behavioral analysis, we record the state of the agent's model (network weights) at different points of training: 5\% and 100\%. To extract an approximate policy of each agent, we manually pick a map instance, shown on figure~\ref{fig:map}, and run a simulation for 5000 steps per trained model and record the agent's current position. We then use heatmaps to visualize the frequency the agent stays at each position. 

We calculate {\em policy divergence} for each fully trained model. We use the {\em no-IM} policy as our baseline behavior. We randomly select 10 map instances and run a simulation on each of trained models for 5000 steps, recording their positions on the grid. 
We calculate policy divergence as $\frac{1}{N} \sum_{i,j} |S_{(i,j)} - S'_{(i,j)}|$, where $N$ is the total number of steps in the simulation (5000), and $S$ and $S'$ are the visitation frequency of the grid position $(i,j)$, for the IM method and the no-IM baseline respectively. Note that $\sum_{i,j} S_{(i,j)} = \sum_{i,j} S'_{(i,j)} = N$. We then average this metric over the 10 simulations. A policy divergence of 0 indicates the two compared policies are equivalent, whereas 2 indicates maximally divergent policies. 

\section{Results}

\subsection{Return Performance}

\begin{figure*}
    \centering
    \includegraphics[width=\linewidth]{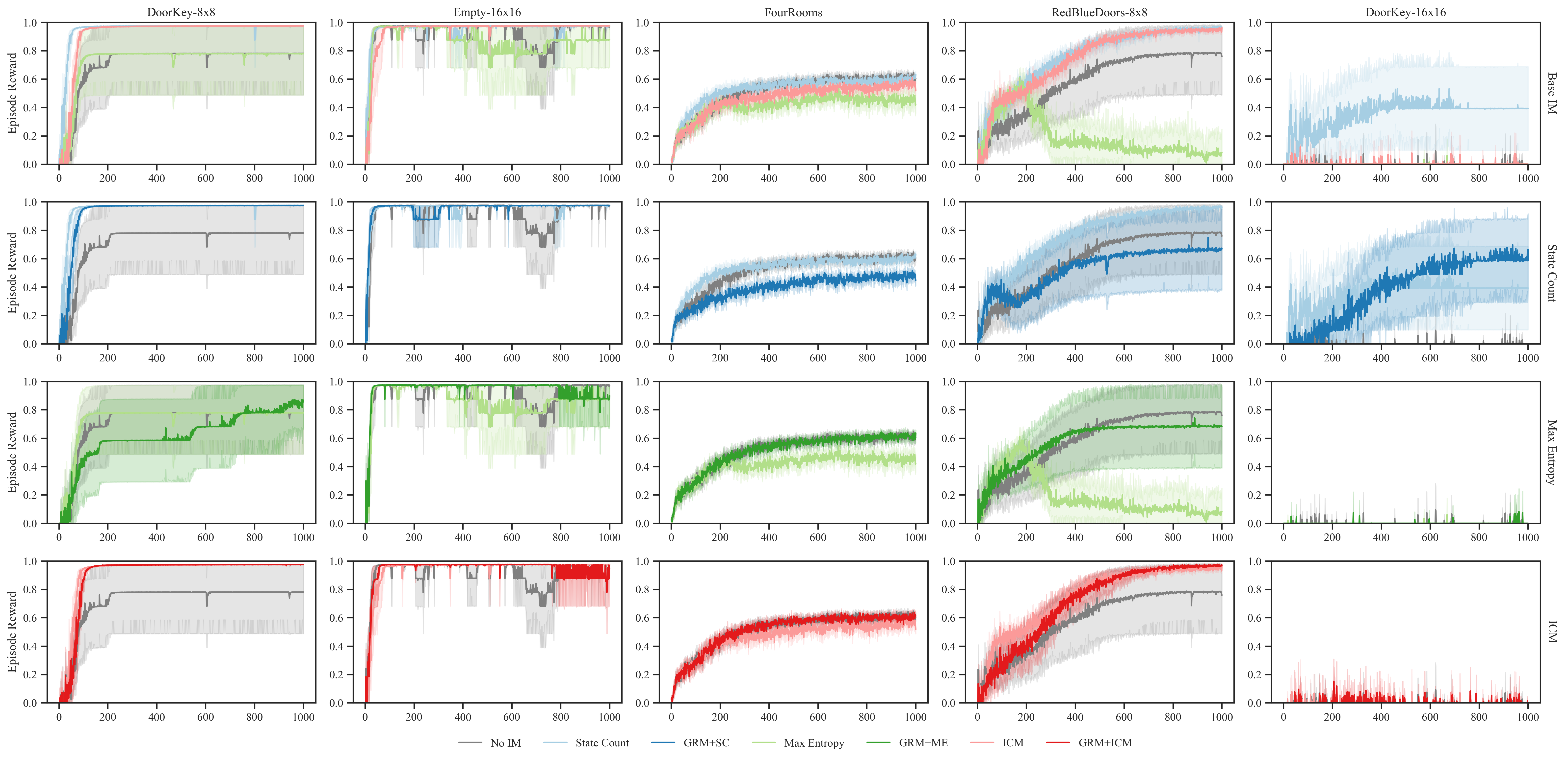}
    \caption{Episodic rewards per iteration of the all trained models. Columns group results by map and rows by type of IM: 1) non-GRM, 2) State Count, 3) Max Entropy, and 4) ICM. Rows 2 and onward include models with and without GRM. Results are averaged over 10 runs, and shading is standard deviation.}
    \label{fig:rewards}
\end{figure*}

\begin{figure*}
    \centering
    \includegraphics[width=\linewidth]{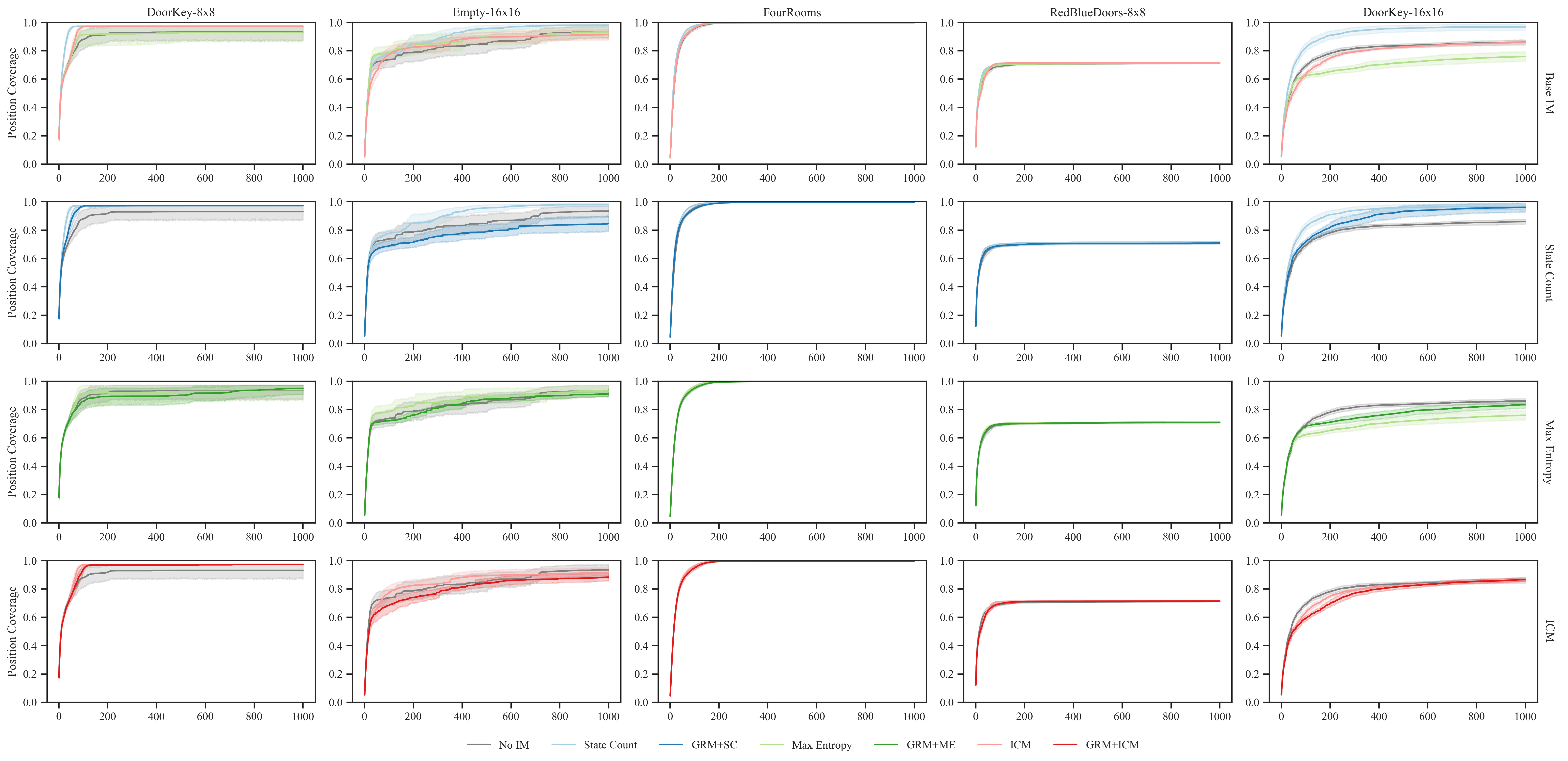}
    \caption{Position (tiles in grid) coverage per iteration of the all trained models. Columns group results by map and rows by type of IM: 1) non-GRM, 2) State Count, 3) Max Entropy, and 4) ICM. Rows 2 and onward include models with and without GRM. Results are averaged over 10 runs, and shading is standard deviation.}
    \label{fig:positions}
\end{figure*}

Figures~\ref{fig:rewards} and ~\ref{fig:positions} respectively show the average episodic rewards and position coverage obtained per by every trained agent, grouped by type of IM. We analyze these results by map.

{\bf DoorKey-8x8: } (first column in all figures) All IM agents result on earlier instances of the maximum reward (around 0.95). No-IM failed on two out of ten runs, resulting on an average final reward of around $0.78$. State Count is the most effective method for earlier convergence, followed by ICM and then Max Entropy. Max Entropy acts similar to no-IM, where it only learns on eight out of ten runs. In this map, GRM makes the agent reach the maximum reward later than using no GRM. In the case of GRM with Max Entropy, the early model performance becomes much worse than plain Max Entropy, although it eventually achieves better performance. For position coverage, ICM and State Count (especially) provide more coverage on the early iterations. GRM methods offer similarly an in-between performance between IM and no-IM. 

{\bf Empty:} (second column in all figures), reward metics follow a similar trend of offering slightly earlier convergence to an optimal policy compared to no-IM, with the exception of both versions of ICM. Notably, GRM methods offer similar or better returns than their respective non-GRM counterparts. Both versions of Max Entropy and GRM+ICM introduce a lot more instability, as catastrophic forgetting happens (for one instance) during the latter stages of training. In terms of coverage, IM methods did offer more position coverage than no-IM. However, this coverage was mostly on the latter stages of training, which is not that useful since the optimal policy has been reached at that point. Conversely, GRM methods actually explore \textit{less} than no-IM. Given that this is a relatively simple map where a lot of exploration is unnecessary this can be seen as an upside for GRM methods.

{\bf FourRooms:} (third column in all figures), 
Episodic rewards were on average lower for every IM agent except for GRM+Max Entropy. State Count and GRM+ICM maintained similar but slightly worse performance compared to no-IM. Coverage metrics were overall similar for all methods, which is a strike against the exploration-focused IM. While in theory more exploration would be beneficial on FourRooms, it is likely intrinsic reward distracted the agent from the real task, and we will visit this point during behavior analysis.

{\bf RedBlueDoors:} (fourth column in all figures), we have another instance of no-IM being unable to consistently learn a policy with optimal episode rewards. This is due to no-IM failing to learn a good policy in two out of the ten trials. Non-GRM StateCount and the two version of ICM can achieve higher rewards than the baseline, as they learn on all ten instances. Max Entropy converges to a poor policy on this map, whereas GRM with State Count and Max Entropy only learn a near-optimal policy in seven out of ten trials. Position coverage metrics were however very similar between all methods.

{\bf DoorKey-16x16:} (fifth column in all figures), only the State Count agents were able to learn the task on some of the trials. Plain State Count converged on four out of ten, and GRM+State Count on six out of ten. Both methods similarly provide better position coverage than no-IM, whereas all other IM methods had similar or worse metrics.


{\bf Overall Observations:} Generally we found that intrinsic motivation alters the return performance of a reinforcement learning agent. Intrinsic motivation can cause a model to converge on extrinsic rewards earlier to the optimal reward. IM increases the likelihood of learning a policy with rewards close to the optimum with varying degrees of success for different IM strategies. State Count and ICM are particularly effective on the majority of the studied experimental treatments, whereas Max Entropy tended to fail on the more complex maps. Combining GRM with either Max Entropy or State Count improves the return performance of the techniques, particularly for Max Entropy. GRM+State Count however sometimes resulted in worse return performance than State Count except performance on the harder DoorKey map over the base State Count. This needs to be explored further in future work on more complex environments.

\subsection{Perceived Behavior}

\begin{figure*}[ht]
    \centering
    \includegraphics[width=\linewidth]{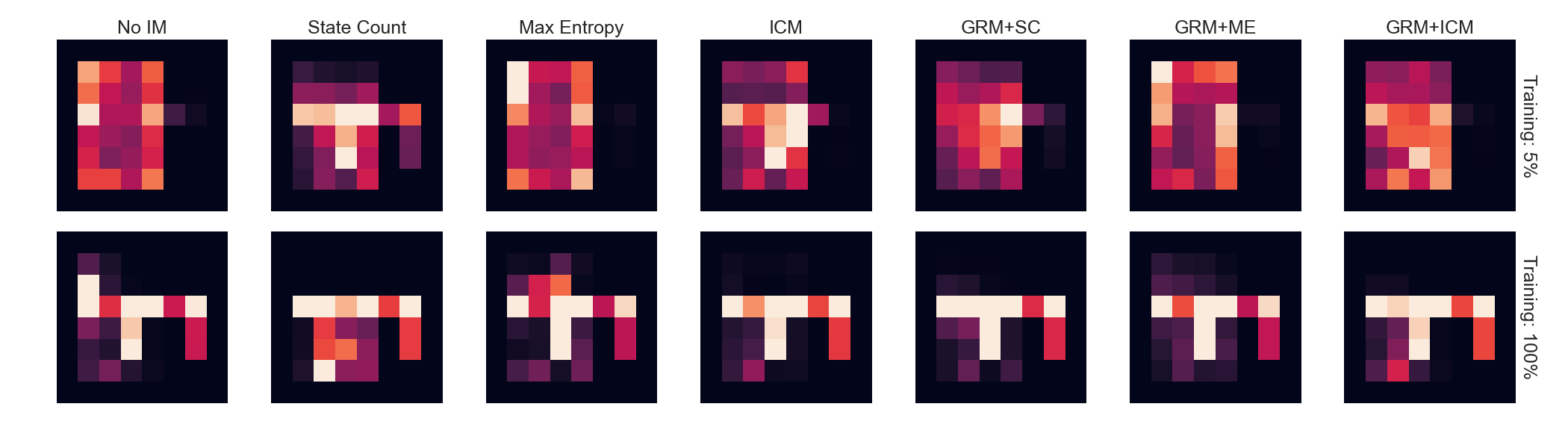}
    \caption{Heatmaps with the position visitation frequency of the seven trained agents through early training and final policy on the DoorKey-8x8 map. Brighter colors indicate a larger fraction of time spent on a grid position.}
    \label{fig:heat-DoorKey-8x8}
\end{figure*}

\begin{figure*}[ht]
    \centering
    \includegraphics[width=\linewidth]{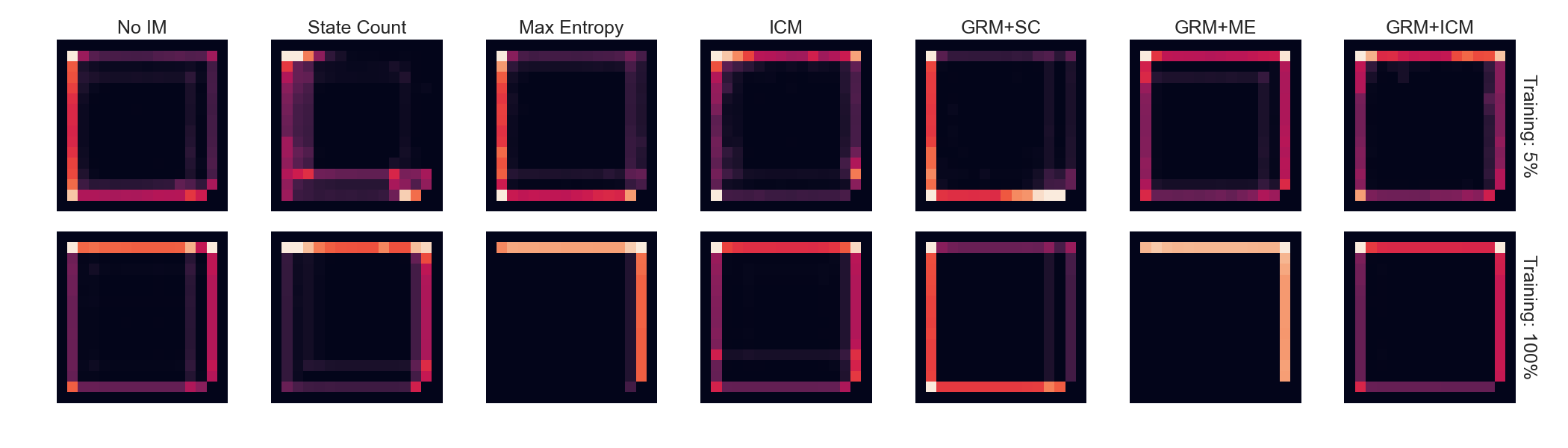}
    \caption{Heatmaps with the position visitation frequency of the seven trained agents through early training and final policy on the Empty-16x16 map. Brighter colors indicate a larger fraction of time spent on a grid position.}
    \label{fig:heat-Empty}
\end{figure*}

\begin{figure*}[ht]
    \centering
    \includegraphics[width=\linewidth]{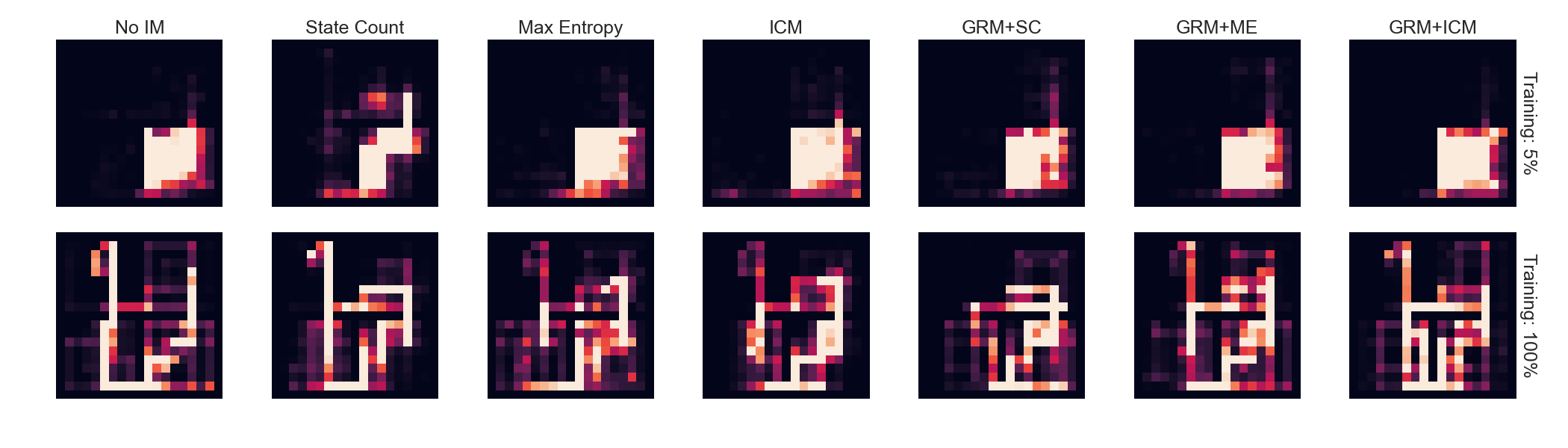}
    \caption{Heatmaps with the position visitation frequency of the seven trained agents through early training and final policy on the FouRooms map. Brighter colors indicate a larger fraction of time spent on a grid position.}
    \label{fig:heat-FourRooms}
\end{figure*}

\begin{figure*}[ht]
    \centering
    \includegraphics[width=\linewidth]{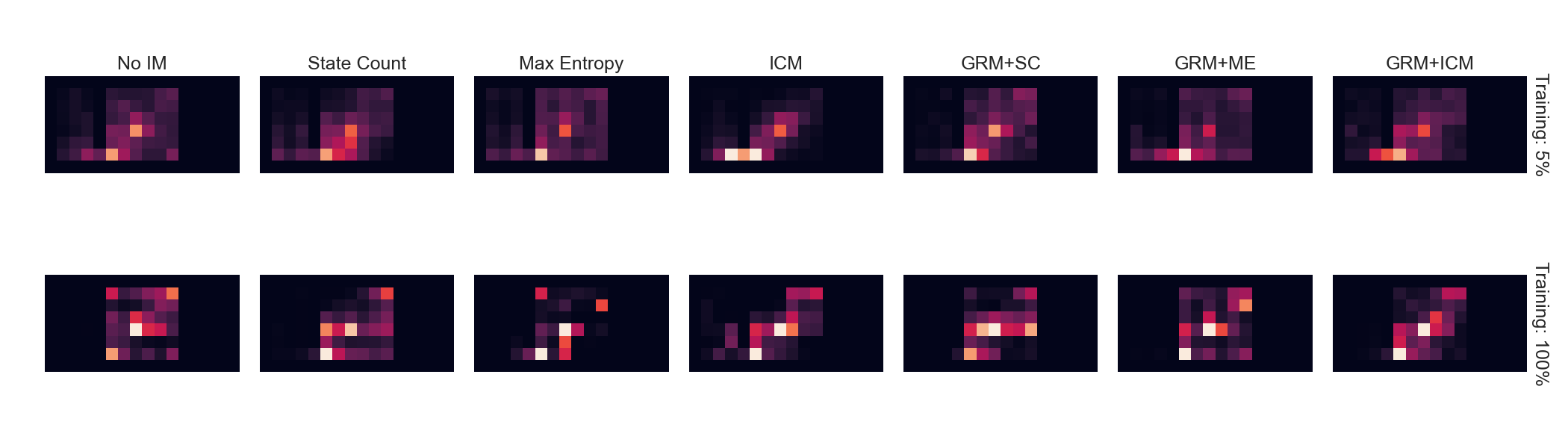}
    \caption{Heatmaps with the position visitation frequency of the seven trained agents through early training and final policy on the RedBlueDoors-8x8 map. Brighter colors indicate a larger fraction of time spent on a grid position.}
    \label{fig:heat-RedBlueDoors}
\end{figure*}

Figures~\ref{fig:heat-DoorKey-8x8}, \ref{fig:heat-Empty}, \ref{fig:heat-FourRooms}, and \ref{fig:heat-RedBlueDoors} show heatmaps of all trained agents (maximum values are clipped for display purposes). We omit DoorKey-16x16 since most models do not converge within the training runs for this experiment. The appropriate map instance is shown figure~\ref{fig:map}. Table~\ref{tab:divergence} shows the average policy divergence for all IM methods.

\begin{table}[ht]
\centering
\caption{Average policy divergence for IM models. Lowest values per map are highlighted.}
\label{tab:divergence}
\begin{tabular}{@{}lrrrr@{}}
\toprule
            & DK            & Empty         & FR            & RBD           \\
\midrule
State Count & 0.40          & 0.42          & \textbf{0.62} & \textbf{0.50} \\
Max Entropy & 0.43          & 0.77          & 0.84          & 0.99          \\
ICM         & \textbf{0.31} & \textbf{0.27} & 0.76          & 0.77          \\
GRM+SC      & 0.37          & 0.79          & 0.72          & 0.61          \\
GRM+ME      & 0.48          & 0.78          & 0.67          & 0.55          \\
GRM+ICM     & 0.49          & 0.32          & 0.74          & 0.63          \\
\bottomrule
\end{tabular}
\end{table}

{\bf DoorKey:}
On the DoorKey-8x8 map~(Figure~\ref{fig:heat-DoorKey-8x8}, map on figure~\ref{fig:map-dk8}), the optimal policy involves the agent going two tiles right, then down for the key, then move towards the door, open it, and finally traverse to the goal.

The baseline (no-IM) agent starts early training by moving somewhat evenly on the left room, favoring the edges. The final policy somewhat reflects our expected behavior, as it it now frequents much more the middle row and the column where the key is on (We will call this the \textit{T}). There is some suboptimal behavior, as the agent sometimes takes a route near the edges of the map to grab the key, as well as moving upwards before approaching they.

The model trained with State Count finds more quickly instances of the sparse reward, which reflects on the earlier focus on the \textit{T} on the left room. The final policy is somewhat unstable however, as it can take multiple paths from the middle to the key: either from the right, the top, or the left, although it seems to mostly favor the left even though it requires one more tile of movement. Its GRM counterpart actually starts out exploring less (thus not having the \textit{T} on the first instances of training), but later develops a much more developed policy that strongly favors the \textit{T}.

For Max Entropy, the initial behavior very much resembles no-IM since only the left room is being explored and no reward has been found by the agent. The method derives a policy that, while strongly suggests efficiency by taking the key from the middle column, also shows some tendency of moving to the center top. When the model is also trained with GRM, the initial behavior remains similar, but towards the end of training it has a well defined policy, where the optimal path is taken the most frequently, similar to GRM with State Count.

In the case of ICM, the model follows an initial behavior that favors the middle of the map, possibly due to finding early instances of the sparse reward or due to the positioning of the key. The method settles for an almost optimal policy towards the end, as it takes the shortest path, with the caveat of sometimes picking up the key from the left side. Adding GRM to this technique does not change much of the behavior for all three snapshots taken, although its final policy favors the position to the left of the key more frequently.

{\em Policy Divergence:} The lowest value was obtained by ICM, which is consistent with our heatmap analysis. While Non-GRM State Count and Max Entropy result on relatively low values, they also engage on behavior that is not optimal, as they still tend to visit the edges of the map towards the end of the training. This can be an effect of reward hacking in action. While GRM+ME shows a higher divergence, the behavioral analysis shows it results on a more `desirable' policy. The IM model resulted on a better policy. GRM+ICM engages on behavior resembling reward hacking more than plain ICM, as it also tends to visit the edges of the map more frequently.

{\bf Empty:}
On the Empty-16x16 map~(Figure~\ref{fig:heat-Empty}, map on figure~\ref{fig:map-empty}), any path that goes towards the bottom right will obtain a good reward ($>$0.9). However, every turn that the agent takes counts as an additional time step, and for this reason the optimal policy is to moves the agent right through topmost row (since the agent starts facing right) and then down the rightmost column.

The baseline (no-IM) policy quickly finds sparse-reward instances of the reward on the early parts of training. The final policy found by no-IM often picks the top path, then often going to the rightmost tile and then down to the goal. Some percentage of the time, however, the agent might go down two tiles before reaching the end (the right-side wall becomes visible at this point). Sometimes the agent takes the bottom path to the goal. As such, the final policy found by the no-IM agent is technically suboptimal.

State Count visits the four leftmost columns and three bottom rows during early training. The final behavior learned by the agent is comparable to no-IM, having two alternate paths on the top side of the map. Similarly it takes the bottom path as well, although less frequently than the baseline, and less often even takes a path that goes towards the middle. Its GRM counterpart behaves much differently, doing less exploration and favoring much more the bottom path. Its final policy resembles much more no-IM, except it takes the bottom path much more frequently.

The initial behavior for the Max Entropy agent is comparable to no-IM, mainly favoring the bottom path and staying near the walls. The final policy is almost perfect, as it only selects the top path. However it can also take the second-to-last column, which is inefficient since it adds to additional turns. The GRM version of the agent behaves similarly, and its final policy follows the optimal behavior.

ICM acts as an in-between State Count and Max Entropy, where it tends to explore the middle areas but still favoring the positions where walls are visible. The final policy ends up behaving more akin to State Count, since it has an additional alternate path should the agent take the bottom side of the map. ICM with GRM is `more stable', as it quickly learns to follow the edges by the middle of the training, and ends with behavior resembling the no-IM policy with no alternate paths.

{\em Policy Divergence:} The methods with the least variance were ICM and GRM+ICM. Max Entropy resulted in the most varied policies. Policies found by IM were better than the one found by no-IM. We also observe that these methods don't really suffer from any common `reward hacking' problems such as visiting tiles on the middle. On the other hand, while State Count offers the added benefits of early exploration, it also results in a variety of paths. GRM with State Count suffers less from this problem, but also results on a slightly worse average policy. In this map, GRM has the effect of making the final policy less similar to using no-IM. However, in the case of ICM and Max Entropy, this is actually better since it results on more desirable behavior; and in the case of State Count, it improves stability.

{\bf FourRooms:}
On the FourRooms map~(Figure~\ref{fig:heat-FourRooms}, map on figure~\ref{fig:map-fr}), an ideal agent with full vision of the map can take to paths to the goal: going through the left room or through the top room. The shortest path in tiles is through the left room. Given the high uncertainty on this map, agents have to balance exploration within rooms against exploration of other rooms.

The baseline model during the early training primarily moves through the initial room, likely due to having few or no instances of the sparse reward. As it learns, the no-IM model starts visiting other rooms as its primary behavior, with some secondary exploration within rooms as it likely searches for the goal tile. By the end of training the agent mainly focuses on moving across rooms, and favors the path through the left room to reach the goal.

State Count changes this initial behavior by emphasizing movement outside of the starting room through the early training, it being the only IM technique that does so. The final policy has established a mostly streamlined path to the goal tile, with some in-room exploration. When trained using GRM, the State Count model is much less successful finding the goal tile. Its final policy did not frequent the space around the goal at all, and instead explores the other three rooms.

The Max Entropy rewards mainly cause the agent to remain on the initial room. The final policy shows more streamlined behavior towards finding the goal, but still with too much exploration within rooms. To contrast, the version trained with GRM frequents the room with the goal tile more often, although it still over-explores all other rooms compared to the baseline.

ICM follows a trend similar to Max Entropy where it spends a lot of time exploring the initial and adjacent rooms, thus seldom finding the room with the goal. Although the final policy is more successful at finding the reward than Max Entropy, it is not as well established as no-IM. With GRM, the IM method seems to favor moving from room to room more, and by the end of training the main routes followed by the agent resemble the baseline.

{\em Policy Divergence:}  Intrinsic rewards might be distracting the agent from finding the goal. This is the case based on the observed behavior of the agents on these maps, and another example of reward hacking in action. Excluding State Count, the policy divergence numbers are lower for GRM methods. GRM mitigated some of the issues of reward hacking.

{\bf RedBlueDoors:}
On the RedBlueDoors~(Figure~\ref{fig:heat-RedBlueDoors}, map on figure~\ref{fig:map-rbd}) map, we expect the fully trained agent to favor the movement from the starting point from the red door, and then to the blue door by following a straight line from one side of the room to the other.

Initially, no-IM explores the middle room evenly, sometimes frequenting the right room (which is unnecessary for getting the sparse reward). The final policy appears to access both doors, but does not establish an efficient route to do so.

State Count initially does suffer from over exploring the right room, but quickly establishes opening the red door is good. By the end of training, it has converged to both doors, although still explores the middle room. GRM helps mitigate both exploration issues.

Max Entropy behaves very similar to the baseline at the very beginning. Max Entropy learns to go to the red door and then to the right side of the room efficiently, yet it appears to ignore the blue door at the top. It is possible that it associates the blue door with high entropy (it can give either high reward or not if the red door is open), and learns to avoid it. This is somewhat fixed by adding GRM, resulting on the model learning to open both doors, but results on erratic movement.

From early training, the model build with ICM has established diagonal movement from the starting point to the red door and then to the blue door, which is reinforced over training. While the final policy is not perfect, it seems as the most well defined of all IM methods. Adding GRM removes the over-exploration of the left room, but adds some instability to the movement in the middle.

{\em Policy Divergence:} The model most resembling no-IM was State Count, which is congruent with the heatmap results. We also observe that GRM greatly reduces divergence for Max Entropy and ICM. In this case the baseline policy is not what we expect to be optimal. Instead, State Count or ICM more closely resemble what would be the ideal behavior.

{\bf Large DoorKey:}
On the DoorKey-16x16 map, the no-IM model is incapable of learning the task. Through training it almost exclusively remains on the bottom left side of the map, and in the end only learns to go upward. All non-State Count IM models follow this trend, although Max Entropy in particular seems to learn to move around the key. State Count methods, the only ones to reach the sparse reward consistently, start by similarly favoring the left and middle sides of the map, but eventually find a consistent horizontal path near the bottom of the map, possibly used to reach the key and/or the door. Since both reach the rightmost states of the map, we can confirm that they have policies that can get to the goal. Visually it seems like the policies have not completely converged and still have some room for improvement.


{\bf Overall Observations:} We found that IM does alter the final policy found by a fair margin. Different sources of IM resulted on varied behavior. State Count does offer the earliest convergence to a functional policy~(i.e. one that can find the sparse reward) but also suffers from over-exploration. Max Entropy is great for stabilizing behavior when the model has access to a sufficient number of sparse-reward instances, but can result on stagnation when a lot of exploration is required. ICM offers a less extreme version of State Count, and often resulted on the less divergent models when little exploration was required. We found that GRM can reduce the policy variance for non-State Count methods as well as consistently produce policies resemble the ideal behavior.

\section{Conclusions and Discussion}
In this paper, we explore the effects of intrinsic motivation techniques through analysis of agent behavior across several variants. As a first step, we empirically analyze how three IM methods, plus GRM, change the behavior of agents trained on MiniGrid, in terms of return performance and observable behavior over training. Results indicate that IM is beneficial in most scenarios but with varied observed in-game behavior of agents. There were both cases where reward hacking made the final policy of IM agents more and less optimal, hinting the side effects of IM rewards might not be as undesirable as the literature suggests. While GRM has the theoretical guarantees of being policy-invariant, results showed it still creates behavior that deviates from the baseline for shorter training runs.

{\bf What is the behavior archetype of the IM methods?} State Count methods result in more exploration during early training, and with good policies. Even during the latter stages of learning, these models have a stronger tendency to move out of the beaten path. Max Entropy on the other hand tends to result on behavior that heavily favors a limited area of the space, resulting on a small number of well traversed (lighter colored in the heatmap) grid positions. We observe Max Entropy by itself has a hard time finding instances of the sparse reward, but it refines the final policy. It results on agents that act very risk averse and prefer the `safe' areas of the map. ICM works as a less extreme version of State Count, offering less early and late exploration.

{\bf Does IM truly result in sub-optimal policies?} The results of this study mainly support that IM methods resulted in policies which we perceived to be closer to optimal. It is likely that longer training would result in the baseline model converging to the expected optimal policy. On the other hand, given enough time it is likely that IM methods that run on diminishing rewards such as State Count will lead to the same result. IM achieved better policies in situations under uncertainty.

{\bf Are theoretical guarantees of optimality enough?} GRM methods with theoretical guarantees on optimality resulted in policies that deviated from the baseline in small training horizons. While theoretically GRM guarantees an invariant policies, empirically there might not be enough time or resources to achieve optimality. 
Selection of an appropriate IM method requires proper characterization of the task complexity and computational resources.

{\bf Is this type of behavior analysis enough?} Much work is to be done in this regard. By including a baseline behavior observation we are one step closer to talking about policy-(in)variance for IM. These results raise some interesting questions. Interpreting the heatmaps is not trivial and disregards the order of operations done by the agent, but switching to empirical video analysis has not been explored yet. Analysis of optimal behavior is harder since the optimal policy changes from map instance to instance, and sometimes the true optimal policy is unknown, especially for complex games. In addition to policy divergence, there may be other, undiscovered behavior-related metrics.

{\bf What are the difference in results with the baseline study} Comparing the grid encoding results from the baseline study to ours we encountered some incongruences. Firstly, our models for DoorKey-16x16 did not converge, which could be due to reduced training time (2M vs. 40M steps). In addition, we generate the heatmaps using models trained across different instances of each map, contrary to their single instance method in the base study. Lastly, the maximum reward for FourRooms was close to 0.9 for the baseline, while our results indicate a lot of instability and rewards closer to 0.6 points. One possible explanation is that the authors of the original study trained on a single instance of the environment per run. 

{\bf What is our recommended intrinsic motivation?} Similar to \cite{kayal_impact_2025} we found State Count to be consistently effective at finding earlier instances of the sparse reward. There is potential for combining State Count with Max Entropy. The former being used during early training and the latter over the second part of the process, perhaps with a coefficient that accordingly decreases/increases. We note that GRM combined with ICM has potential, since it tended to `smooth' over the final policy while retaining similar early exploration.

\section{Threats to Validity}
We identified a group of factors that could have potentially influenced the results of the experiment. We used a static set of hyperparameters for PPO, which we based on ~\cite{kayal_impact_2025}. We adjusted the value for the intrinsic reward coefficient $\beta$ individually, but non-exhaustively. The implementation details of the PPO algorithm can affect its performance\footnote{See https://iclr-blog-track.github.io/2022/03/25/ppo-implementation-details/.}, which we mitigate by using the existing DEIR implementation.

{\bf Training Data:} The models are trained for 10 million frames total, which we found enough to find a stable policy in most cases. However, the possibility remains that with a longer training period the performance of the models might change. We repeat the experiments for a total of 10 times, which is double as many runs as the protocol study.

{\bf Metrics:} For our return performance analysis, we used the metrics proposed in~\cite{kayal_impact_2025}. For our behavior analysis however we rely on heatmap and visual interpretation. To the best of our knowledge, there are no existing methods to analyze the behavior of reinforcement learning agents. In general we report that behavior analysis is an under studied area of RL.

{\bf Generalizability:} Grid-live environments of around the same complexity. Ideally we would work with game environments such as Atari or MicroRTS, but instead chose MiniGrid as we expected behavioral analysis to be too tough of a starting point in those games. We will move on to actual game environments in our following studies.

{\bf Reproducibility:} We have reported technical details of the experiment in the attempt of making these results replicable, and we have shared our source code and artifacts.


\bibliography{aaai24}

\begin{thebibliography}{30}
\providecommand{\natexlab}[1]{#1}

\bibitem[{Andres, Villar-Rodriguez, and Del~Ser(2022)}]{andres2022evaluation}
Andres, A.; Villar-Rodriguez, E.; and Del~Ser, J. 2022.
\newblock An evaluation study of intrinsic motivation techniques applied to reinforcement learning over hard exploration environments.
\newblock In \emph{International Cross-Domain Conference for Machine Learning and Knowledge Extraction}, 201--220. Springer.

\bibitem[{Badia et~al.(2020)Badia, Sprechmann, Vitvitskyi, Guo, Piot, Kapturowski, Tieleman, Arjovsky, Pritzel, Bolt, and Blundell}]{badia2020never}
Badia, A.~P.; Sprechmann, P.; Vitvitskyi, A.; Guo, D.; Piot, B.; Kapturowski, S.; Tieleman, O.; Arjovsky, M.; Pritzel, A.; Bolt, A.; and Blundell, C. 2020.
\newblock Never Give Up: Learning Directed Exploration Strategies.
\newblock In \emph{International Conference on Learning Representations}.

\bibitem[{Behboudian et~al.(2022)Behboudian, Satsangi, Taylor, Harutyunyan, and Bowling}]{behboudian_policy_2022}
Behboudian, P.; Satsangi, Y.; Taylor, M.~E.; Harutyunyan, A.; and Bowling, M. 2022.
\newblock Policy invariant explicit shaping: an efficient alternative to reward shaping.
\newblock \emph{Neural Computing and Applications}, 1--14.

\bibitem[{Burda et~al.(2018{\natexlab{a}})Burda, Edwards, Pathak, Storkey, Darrell, and Efros}]{burda_large-scale_2018}
Burda, Y.; Edwards, H.; Pathak, D.; Storkey, A.; Darrell, T.; and Efros, A.~A. 2018{\natexlab{a}}.
\newblock Large-scale study of curiosity-driven learning.
\newblock \emph{arXiv preprint arXiv:1808.04355}.

\bibitem[{Burda et~al.(2018{\natexlab{b}})Burda, Edwards, Storkey, and Klimov}]{burda_exploration_2018}
Burda, Y.; Edwards, H.; Storkey, A.; and Klimov, O. 2018{\natexlab{b}}.
\newblock Exploration by random network distillation.
\newblock \emph{arXiv preprint arXiv:1810.12894}.

\bibitem[{Chen et~al.(2022)Chen, Hong, Pajarinen, and Agrawal}]{chen2022redeeming}
Chen, E.; Hong, Z.-W.; Pajarinen, J.; and Agrawal, P. 2022.
\newblock Redeeming intrinsic rewards via constrained optimization.
\newblock In Koyejo, S.; Mohamed, S.; Agarwal, A.; Belgrave, D.; Cho, K.; and Oh, A., eds., \emph{Advances in Neural Information Processing Systems}, volume~35, 4996--5008. Curran Associates, Inc.

\bibitem[{Chevalier-Boisvert et~al.(2023)Chevalier-Boisvert, Dai, Towers, Perez-Vicente, Willems, Lahlou, Pal, Castro, and Terry}]{chevalier-boisvert_minigrid_2023}
Chevalier-Boisvert, M.; Dai, B.; Towers, M.; Perez-Vicente, R.; Willems, L.; Lahlou, S.; Pal, S.; Castro, P.~S.; and Terry, J. 2023.
\newblock Minigrid \& {Miniworld}: {Modular} \& {Customizable} {Reinforcement} {Learning} {Environments} for {Goal}-{Oriented} {Tasks}.
\newblock \emph{Advances in Neural Information Processing Systems}, 36: 73383--73394.

\bibitem[{Colas et~al.(2022)Colas, Karch, Sigaud, and Oudeyer}]{colas2022autotelic}
Colas, C.; Karch, T.; Sigaud, O.; and Oudeyer, P.-Y. 2022.
\newblock Autotelic agents with intrinsically motivated goal-conditioned reinforcement learning: a short survey.
\newblock \emph{Journal of Artificial Intelligence Research}, 74: 1159--1199.

\bibitem[{Forbes et~al.(2024{\natexlab{a}})Forbes, Gupta, Villalobos-Arias, Potts, Jhala, and Roberts}]{forbes2024potential}
Forbes, G.~C.; Gupta, N.; Villalobos-Arias, L.; Potts, C.~M.; Jhala, A.; and Roberts, D.~L. 2024{\natexlab{a}}.
\newblock Potential-Based Reward Shaping for Intrinsic Motivation.
\newblock In \emph{Proceedings of the 23rd International Conference on Autonomous Agents and Multiagent Systems}, 589--597.

\bibitem[{Forbes et~al.(2024{\natexlab{b}})Forbes, Villalobos-Arias, Wang, Jhala, and Roberts}]{forbes_potential-based_2024}
Forbes, G.~C.; Villalobos-Arias, L.; Wang, J.; Jhala, A.; and Roberts, D.~L. 2024{\natexlab{b}}.
\newblock Potential-Based Intrinsic Motivation: Preserving Optimality With Complex, Non-Markovian Shaping Rewards.
\newblock \emph{arXiv preprint arXiv:2410.12197}.

\bibitem[{Forbes et~al.(2025)Forbes, Wang, Villalobos-Arias, Jhala, and Roberts}]{forbes2025action}
Forbes, G.~C.; Wang, J.; Villalobos-Arias, L.; Jhala, A.; and Roberts, D.~L. 2025.
\newblock Action-Dependent Optimality-Preserving Reward Shaping.
\newblock \emph{arXiv preprint arXiv:2505.12611}.

\bibitem[{Huang and Ontañón(2020)}]{huang_action_2020}
Huang, S.; and Ontañón, S. 2020.
\newblock Action {Guidance}: {Getting} the {Best} of {Sparse} {Rewards} and {Shaped} {Rewards} for {Real}-time {Strategy} {Games}.
\newblock ArXiv:2010.03956 [cs].

\bibitem[{Jordan et~al.(2024)Jordan, White, Silva, White, and Thomas}]{jordan2024position}
Jordan, S.~M.; White, A.; Silva, B. C.~D.; White, M.; and Thomas, P.~S. 2024.
\newblock Position: Benchmarking is Limited in Reinforcement Learning Research.
\newblock In Salakhutdinov, R.; Kolter, Z.; Heller, K.; Weller, A.; Oliver, N.; Scarlett, J.; and Berkenkamp, F., eds., \emph{Proceedings of the 41st International Conference on Machine Learning}, volume 235 of \emph{Proceedings of Machine Learning Research}, 22551--22569. PMLR.

\bibitem[{Kayal, Pignatelli, and Toni(2025)}]{kayal_impact_2025}
Kayal, A.; Pignatelli, E.; and Toni, L. 2025.
\newblock The impact of intrinsic rewards on exploration in {Reinforcement} {Learning}.
\newblock ArXiv:2501.11533 [cs].

\bibitem[{Laskin et~al.(2021)Laskin, Yarats, Liu, Lee, Zhan, Lu, Cang, Pinto, and Abbeel}]{laskin2021urlb}
Laskin, M.; Yarats, D.; Liu, H.; Lee, K.; Zhan, A.; Lu, K.; Cang, C.; Pinto, L.; and Abbeel, P. 2021.
\newblock {URLB}: Unsupervised Reinforcement Learning Benchmark.
\newblock In \emph{Thirty-fifth Conference on Neural Information Processing Systems Datasets and Benchmarks Track (Round 2)}.

\bibitem[{Le et~al.(2024)Le, Do, Nguyen, and Venkatesh}]{le_beyond_2024}
Le, H.; Do, K.; Nguyen, D.; and Venkatesh, S. 2024.
\newblock Beyond {Surprise}: {Improving} {Exploration} {Through} {Surprise} {Novelty}.
\newblock In \emph{Proceedings of the 23rd {International} {Conference} on {Autonomous} {Agents} and {Multiagent} {Systems}}, 1084--1092.

\bibitem[{Liu, Gu, and Liu(2019)}]{liu2019policy}
Liu, J.; Gu, X.; and Liu, S. 2019.
\newblock Policy optimization reinforcement learning with entropy regularization.
\newblock \emph{arXiv preprint arXiv:1912.01557}.

\bibitem[{Mnih et~al.(2013)Mnih, Kavukcuoglu, Silver, Graves, Antonoglou, Wierstra, and Riedmiller}]{mnih2013playing}
Mnih, V.; Kavukcuoglu, K.; Silver, D.; Graves, A.; Antonoglou, I.; Wierstra, D.; and Riedmiller, M. 2013.
\newblock Playing atari with deep reinforcement learning.
\newblock \emph{arXiv preprint arXiv:1312.5602}.

\bibitem[{Mnih et~al.(2015)Mnih, Kavukcuoglu, Silver, Rusu, Veness, Bellemare, Graves, Riedmiller, Fidjeland, Ostrovski et~al.}]{mnih2015human}
Mnih, V.; Kavukcuoglu, K.; Silver, D.; Rusu, A.~A.; Veness, J.; Bellemare, M.~G.; Graves, A.; Riedmiller, M.; Fidjeland, A.~K.; Ostrovski, G.; et~al. 2015.
\newblock Human-level control through deep reinforcement learning.
\newblock \emph{nature}, 518(7540): 529--533.

\bibitem[{Mohamed and Jimenez~Rezende(2015)}]{mohamed2015variational}
Mohamed, S.; and Jimenez~Rezende, D. 2015.
\newblock Variational information maximisation for intrinsically motivated reinforcement learning.
\newblock \emph{Advances in neural information processing systems}, 28.

\bibitem[{Onta{\~n}{\'o}n et~al.(2018)Onta{\~n}{\'o}n, Barriga, Silva, Moraes, and Lelis}]{ontanon2018first}
Onta{\~n}{\'o}n, S.; Barriga, N.~A.; Silva, C.~R.; Moraes, R.~O.; and Lelis, L.~H. 2018.
\newblock The first microrts artificial intelligence competition.
\newblock \emph{AI Magazine}, 39(1): 75--83.

\bibitem[{Oudeyer and Kaplan(2007)}]{oudeyer2007intrinsic}
Oudeyer, P.-Y.; and Kaplan, F. 2007.
\newblock What is intrinsic motivation? A typology of computational approaches.
\newblock \emph{Frontiers in neurorobotics}, 1: 108.

\bibitem[{Pathak et~al.(2017)Pathak, Agrawal, Efros, and Darrell}]{pathak_curiosity-driven_2017}
Pathak, D.; Agrawal, P.; Efros, A.~A.; and Darrell, T. 2017.
\newblock Curiosity-driven exploration by self-supervised prediction.
\newblock In \emph{International conference on machine learning}, 2778--2787. PMLR.

\bibitem[{Raileanu and Rocktäschel(2020)}]{raileanu2020ride}
Raileanu, R.; and Rocktäschel, T. 2020.
\newblock RIDE: Rewarding Impact-Driven Exploration for Procedurally-Generated Environments.
\newblock In \emph{International Conference on Learning Representations}.

\bibitem[{Schulman et~al.(2017)Schulman, Wolski, Dhariwal, Radford, and Klimov}]{schulman_proximal_2017}
Schulman, J.; Wolski, F.; Dhariwal, P.; Radford, A.; and Klimov, O. 2017.
\newblock Proximal policy optimization algorithms.
\newblock \emph{arXiv preprint arXiv:1707.06347}.

\bibitem[{Strehl and Littman(2008)}]{strehl2008analysis}
Strehl, A.~L.; and Littman, M.~L. 2008.
\newblock An analysis of model-based interval estimation for Markov decision processes.
\newblock \emph{Journal of Computer and System Sciences}, 74(8): 1309--1331.

\bibitem[{Taiga et~al.(2020)Taiga, Fedus, Machado, Courville, and Bellemare}]{taiga2020bonus}
Taiga, A.~A.; Fedus, W.; Machado, M.~C.; Courville, A.; and Bellemare, M.~G. 2020.
\newblock On Bonus Based Exploration Methods In The Arcade Learning Environment.
\newblock In \emph{International Conference on Learning Representations}.

\bibitem[{Vinyals et~al.(2019)Vinyals, Babuschkin, Czarnecki, Mathieu, Dudzik, Chung, Choi, Powell, Ewalds, Georgiev et~al.}]{vinyals2019grandmaster}
Vinyals, O.; Babuschkin, I.; Czarnecki, W.~M.; Mathieu, M.; Dudzik, A.; Chung, J.; Choi, D.~H.; Powell, R.; Ewalds, T.; Georgiev, P.; et~al. 2019.
\newblock Grandmaster level in StarCraft II using multi-agent reinforcement learning.
\newblock \emph{nature}, 575(7782): 350--354.

\bibitem[{Wan et~al.(2023)Wan, Tang, Tian, and Kaneko}]{wan2023deir}
Wan, S.; Tang, Y.; Tian, Y.; and Kaneko, T. 2023.
\newblock DEIR: Efficient and Robust Exploration through Discriminative-Model-Based Episodic Intrinsic Rewards.
\newblock In Elkind, E., ed., \emph{Proceedings of the Thirty-Second International Joint Conference on Artificial Intelligence, {IJCAI-23}}, 4289--4298. International Joint Conferences on Artificial Intelligence Organization.
\newblock Main Track.

\bibitem[{Zhang et~al.(2021)Zhang, Xu, Wang, Wu, Keutzer, Gonzalez, and Tian}]{zhang2021noveld}
Zhang, T.; Xu, H.; Wang, X.; Wu, Y.; Keutzer, K.; Gonzalez, J.~E.; and Tian, Y. 2021.
\newblock Noveld: A simple yet effective exploration criterion.
\newblock \emph{Advances in Neural Information Processing Systems}, 34: 25217--25230.

\end{thebibliography}

\end{document}